\documentclass[letterpaper, 10 pt, conference]{ieeeconf}  

\IEEEoverridecommandlockouts                              

\usepackage{cite}
\usepackage{amsmath,amssymb,amsfonts}
\usepackage{graphicx}
\usepackage{float}
\usepackage{textcomp}
\usepackage{xcolor}
\usepackage{algorithm}
\usepackage{algorithmic}
\usepackage{url}
\usepackage{nomencl}
\usepackage{siunitx}
\usepackage[font=small]{caption}

\usepackage[colorlinks=true,linkcolor=blue, urlcolor=black, citecolor=cyan]{hyperref}

\def\BibTeX{{\rm B\kern-.05em{\sc i\kern-.025em b}\kern-.08em
    T\kern-.1667em\lower.7ex\hbox{E}\kern-.125emX}}

\title{\LARGE \bf
MorphoMove: Bi-Modal Path Planner with MPC-based Path Follower for Multi-Limb Morphogenetic UAV
}

\author{Muhammad Ahsan Mustafa, Yasheerah Yaqoot, Mikhail Martynov, Sausar Karaf and Dzmitry Tsetserukou
\thanks{The authors are with the Intelligent Space Robotics Laboratory, CDE, Skolkovo Institute of Science and Technology (Skoltech), Moscow, Russia. {\tt \{Ahsan.Mustafa, Yasheerah.Yaqoot, Mikhail.Martynov, Sausar.Karaf, D.Tsetserukou\}@skoltech.ru}}%
}

\begin{document}

\maketitle

\begin{abstract}
This paper discusses developments for a multi-limb morphogenetic UAV, MorphoGear, that is capable of both aerial flight and ground locomotion. A hybrid path planning algorithm based on the A* strategy has been developed, enabling seamless transition between air-to-ground navigation modes, thereby enhancing robot's mobility in complex environments. Moreover, precise path following is achieved during ground locomotion with a Model Predictive Control (MPC) architecture for its novel walking behaviour. Experimental validation was conducted in the Unity simulation environment utilizing Python scripts to compute control values. The algorithm's performance is validated by the Root Mean Squared Error (RMSE) of 0.91 cm and a maximum error of 1.85 cm, as demonstrated by the results. These developments highlight the adaptability of MorphoGear in navigation through cluttered environments, establishing it as a usable tool in autonomous exploration, both aerial and ground-based.

\end{abstract}
\section{Introduction}
In the past decade, there has been a notable surge in the adoption of unmanned ground vehicles (UGVs) and unmanned aerial vehicles (UAVs) across multiple disciplines. These vehicles provide a certain degree of autonomy in terms of various tasks such as, autonomous inspections and monitoring of construction sites, exploration of rough terrains, data collection from complex environments, and reconnaissance \cite{b1}. Each of the fore mentioned vehicles has its own set of drawbacks that limit their real world applicability. These range from overcoming obstacles to limited mobility for the UGVs and limited payloads and power supply for the UAVs. To address this issue, authors \cite{b1} designed and developed a novel morphogenetic UAV, MorphoGear (Fig.~\ref{morphogear}), which integrates both aerial flight and terrestrial locomotion functionalities into a single platform. 

To extend MorphoGear's functionality, seamless transition between aerial and ground navigation is essential. Moreover, it is important that the drone follows the required path with minimum deviations. Path planning for such a robot still remains a challenging task. There needs to be a combination of path planning and path following algorithms that ensures motion with the least errors.

\begin{figure}[t]
    \flushright
    \includegraphics[width=0.48\textwidth]{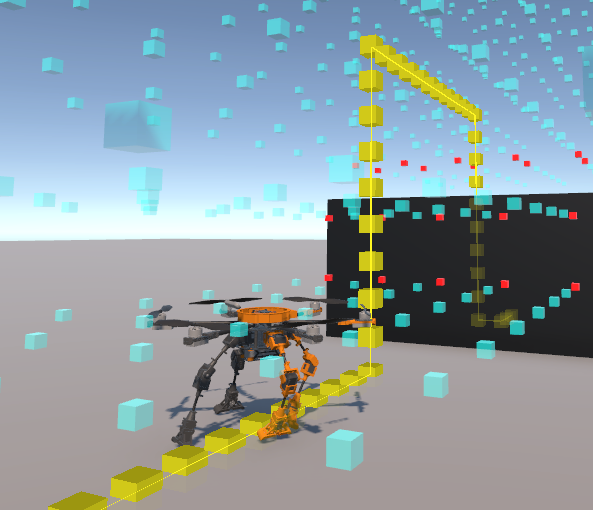}
    \caption{MorphoGear in its mid-step while following the planned trajectory (yellow line) using the designed MPC in Unity environment.}
    \label{main_pic}
\end{figure}

This paper introduces a hybrid path planning algorithm for transition between air/ground modes using the A* technique of cost heuristic function for locomotion and flight modes, and a path following algorithm using Model Predictive Control (MPC) for optimal performance during MorphoGear's ground locomotion (Fig.~\ref{main_pic}) since MPC frameworks for dynamic locomotion prove to be essential for respecting the dynamic constraints of the system \cite{b2}. These algorithms together present a novel approach in the domain of multi-modal motion that enhances MorphoGear's adaptability and efficiency, thus establishing it as a usable tool in autonomous exploration. 

\section{Related Works}
In robotics, path planning and precise path following are important for the efficient and safe navigation of autonomous systems. The advancements in the morphogenetic robotic systems, which have the ability for both ground and aerial locomotion, have introduced new challenges for the conventional path planning methods. 

Graph-search algorithms, like A* \cite{b3} and its variants, are widely used for path planning in 2D and 3D environments \cite{b4}. For dynamic environments where obstacles are unknown or there is not enough information about the environment, D* Lite \cite{b5} serves as a good option by providing incremental search and being memory efficient at the same time; and when the graph changes over time, LPA* \cite{b6} provides efficient re-planning by updating the path finding information. The mentioned variants of the traditional A* algorithm provide optimal paths depending on the environment. 

The path planning algorithms for UGVs and UAVs are different; therefore, they cannot be directly applied to morphogenetic robots. The path planning algorithms for such robots are still in the initial stages of development since the air-to-ground transition brings new challenges in this domain. Several researchers have tried to create a connection between the UAVs and UGVs. Li et al. presented a hybrid path planning method to enhance cooperation between UGVs and UAVs for optimal paths in dynamic environments \cite{b7}. For hybrid robots, Sharif et al. proposed an energy efficient path planning algorithm for a hybrid fly-drive robot that uses an A* algorithm with adjusted heuristic function or additional cost factors related to energy usage into the algorithm's evaluation of potential paths \cite{b8}. Wang et al. introduced a novel approach to planning paths for air-ground robots, considering modal switching points for optimization, thus optimizing paths by dynamically switching between different modes of transportation, such as aerial and ground-based movement \cite{b9}. 

The above mentioned works do not consider any control system for tracking the planned trajectories. Techniques like cascaded PID controllers \cite{b10} are used where the inner loop PID tracks a desired trajectory for the foot's position by controlling the robot dog's leg joint angles while the outer loop regulates the target trajectory itself. Other variations of PID controllers, such as fuzzy PID \cite{b11} are used where fuzzy logic is integrated with traditional PID. Dongho \cite{b12} designed a more complex controller where model predictive control was used in combination with whole body control. 

Most research done in the domain of trajectory tracking for quadrupedal robots considers it to be in a robot-dog-like manner. In this study, due to MorphoGear's novel design and gait manner, a new control strategy leveraging MPC was developed, which was implemented using Python CasADi \cite{b13} interfaces similar to \cite{b14}.

\section{System Overview}
\subsection{MorphoGear UAV}
\begin{figure}[!h]
\centering
\includegraphics[width=0.8\linewidth]{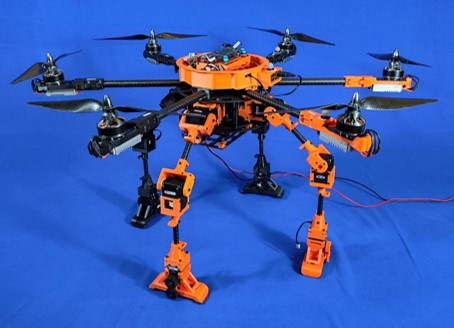} 
\caption{MorphoGear - a multi-limb morphogenetic UAV for aerial and ground locomotion \cite{b1}.} 
\label{morphogear}
\end{figure}

MorphoGear UAV is a robotic drone consisting of a hexacopter with four integrated limbs, designed to locomote and manipulate objects. Each limb has 3 degrees of freedom with a two-finger gripper attached at the end of each limb for object manipulation. Each limb is placed at 90 deg. intervals around the circular base, which results in a fully symmetric design as seen in Fig.~\ref{morphogear}. 

\subsection{Limb Structure of MorphoGear}
\begin{figure}[htbp]
\centering
\includegraphics[width=0.65\linewidth]{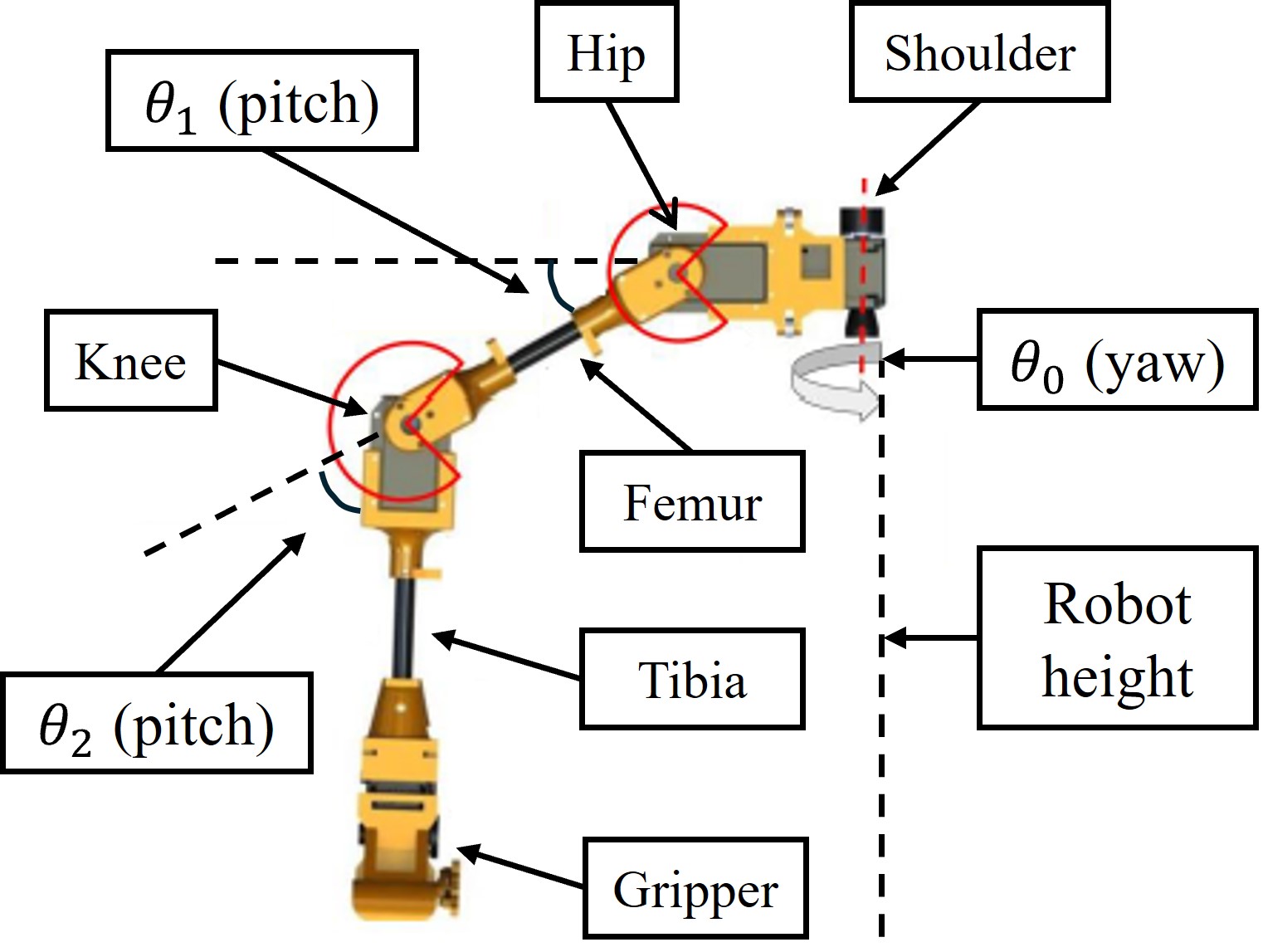} 
\caption{3D CAD depicting the limb structure of MorphoGear \cite{b1}.}
\label{limb_structure}
\end{figure}

This section will provide a brief description focusing on the limbs of MorphoGear, which are controlled by the designed MPC for ground locomotion. The shoulder joint is responsible for the yaw motion of each limb, while the hip and knee joint are responsible for the pitch of the femur and tibia, respectively. The limb structure can be visualised in Fig.~\ref{limb_structure} while its parameters are listed in Table~\ref{tab:MGparams}.
\begin{table}[h!]
    \centering
    \caption{Limb Parameters}
    \label{tab:MGparams}
    \begin{tabular}{|l|c|}
    \hline
    Parameter & Value \\
    \hline
    Robot height, H & 31 cm\\
    Femur length, \( l_F \) & 15.4 cm \\
    Tibia length, \( l_T \) & 20.6 cm\\
    Yaw angle, \( \theta_0 \) & \SI{-35}{\degree} to \SI{+35}{\degree} \\
    Pitch angle, \( \theta_1 \) & \SI{0}{\degree} to \SI{-90}{\degree} \\
    Pitch angle, \( \theta_2 \) & \SI{0}{\degree} to \SI{-90}{\degree} \\
    \hline
    \end{tabular}
\end{table}
A detailed overview of the robot and its mechanics and electronics parts was described in a previous study \cite{b1} and \cite{b15}.

\section{Path Planning}
Recent research regarding heterogeneous swarms with a leader drone, MorphoGear \cite{b16} showed good potential for autonomous inspection.
Reinforcement Learning-based approach to swarm of mini-drones landing on MorphoGear was done in the next work \cite{b17}. However, previously the robot was operated manually. To provide missions autonomously, there is a need to develop a path planner that will be executed accurately without human influence.

The developed path planning algorithm differs from existing ones. This is because MorphoGear has the ability to both fly and walk, therefore, a hybrid path planning algorithm is developed that takes into account this novelty and hence gives a path that uses both walking and flying modes depending on the environment.

\subsection{Bi-Modal Path Planning Algorithm} 
A* algorithm is used as a baseline because it is goal-oriented and easy to implement. The algorithm uses A* with some slight modifications to find a path. The path depends on the environment and the obstacles present in it.

The start and goal positions are taken and mapped into the grid frame. These points along with the grid are taken, and the path is searched in a 2D frame. This is due to the fact that if there is a path in the environment between the start and goal positions that can be found in the 2D plane, then the MorphoGear prefers to use it via ground locomotion. The aerial flight of MorphoGear takes place when there is no path in the 2D plane. This is where the second phase of the algorithm comes in and searches for the path in 3D. But, in order to find the optimal path, it considers the point closest to the goal position in the 2D path using a Euclidean distance function and considers it as an initial point for 3D search. At each iteration of path finding in 3D, the algorithm checks for the possibility to land, i.e., move down in the vertical axis~($Z$). If there is a possibility to go down in $Z$, it is bound to go down, and then find the path to the goal. In short, it prefers walking from the start position to the goal position and flies only when walking is not possible, i.e., in the presence of a wall or barrier. The process is described in Algorithm~\ref{alg:path_plan}.
\begin{algorithm}
\caption{Bi-Modal Path Planning Algorithm}\label{alg:path_plan}
\begin{algorithmic}[1]
\REQUIRE grid, start position, goal position
\ENSURE path
\STATE Initialization for 2D search
\STATE Add start position to open list
\STATE found\_dest $\gets$ False
\WHILE{open list not empty}
    \STATE Perform A* search for 2D
    \IF{destination found}
        \STATE found\_dest $\gets$ True
        \RETURN path
    \ENDIF
\ENDWHILE
\IF{found\_dest == False }
    \FOR{each point in path}
        \STATE x, y, z = point
        \STATE $x_{\text{dest}}, y_{\text{dest}}, z_{\text{dest}}$ = goal position
        \STATE $dist = \sqrt{(x - x_{\text{dest}})^2 + (y - y_{\text{dest}})^2 + (z - z_{\text{dest}})^2}$
        \STATE nearest\_point $\gets$ point with minimum $dist$
    \ENDFOR
    \STATE $last\_point \gets$ nearest\_point
    \STATE Initialization for 3D search
    \STATE Add $last\_point$ to open list
    \WHILE{open list not empty}
        \STATE Perform A* search for 3D
        \IF{destination found}
            \STATE found\_dest $\gets$ True
            \RETURN path
        \ENDIF
        \IF{possibility to move downwards}
            \STATE Add downward position to open list
            \STATE \textbf{break}
        \ENDIF
    \ENDWHILE
\ENDIF
\end{algorithmic}
\end{algorithm}

\begin{figure}[htbp]
\centering
\includegraphics[width=0.9\linewidth]{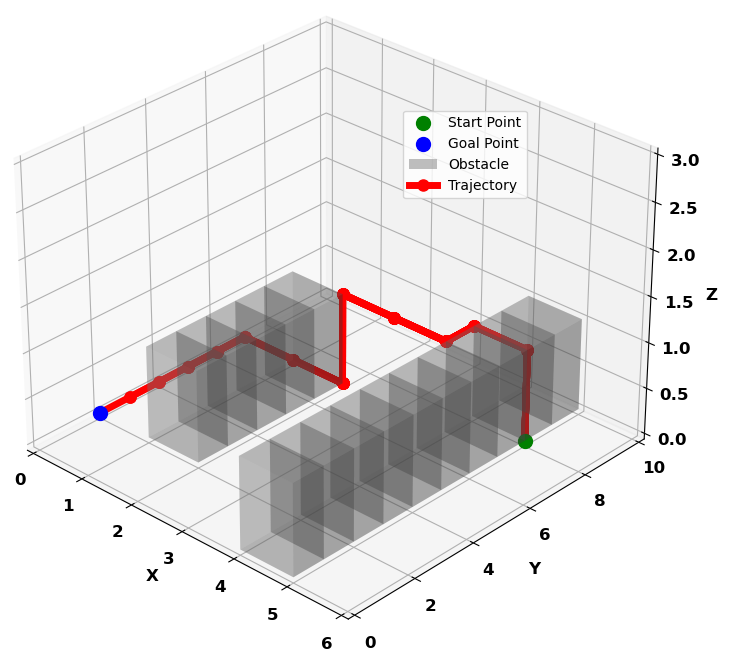} 
\caption{The planned trajectory (red line) of MorphoGear incorporating both aerial and ground paths based on the environment.}
\label{path}
\end{figure}

The path generated using Algorithm~\ref{alg:path_plan} can be seen in Fig.~\ref{path}. It shows one of the configurations where the start position is in front of the barrier, which can only be crossed by flying, and once the barrier is crossed, the algorithm finds the possibility to land and then finds the trajectory in the plane to reach the goal position.

\section{Path Following}
\subsection{Canter Gait Locomotion}\label{subsec:motion}
\begin{figure}[htbp]
\centering
\includegraphics[width=0.9\linewidth]{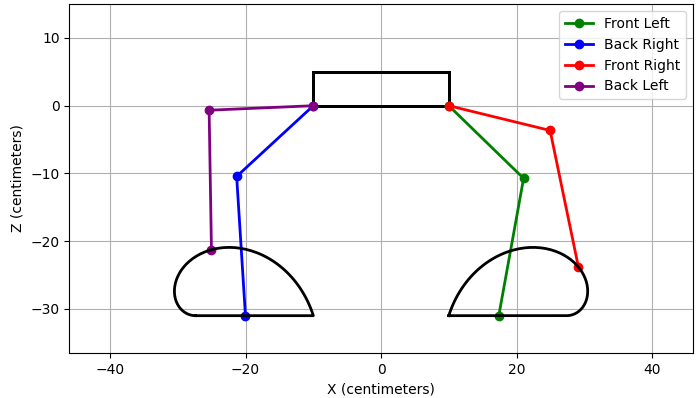} 
\caption{Limb trajectories for canter gait motion to be followed by the end-effector of each limb.}
\label{leg_traj}
\end{figure}

MorphoGear adopts a canter gait for ground locomotion. The concept underlying this gait involves moving all 4 limbs synchronously along an annular trajectory where the top part of it is a section of the Archimedes spiral while the bottom is a straight, flat line. Due to the robot's symmetry, the trajectory for each leg is identical but in opposite directions for the front and rear limbs \cite{b1}. The limb trajectories are shown in Fig.~\ref{leg_traj}.

The inverse kinematic equations explained in \cite{b1} are mentioned below with the parameters listed in Table~\ref{tab:MGparams}.
\begin{equation}
\cos \theta_2 = \frac{{x_2^2 + z_2^2 - l_F^2 - l_T^2}}{{2 l_F l_T}},
\label{eq:IK1}
\end{equation}
\begin{equation}
\theta_2 = -\arccos(\cos(\theta_2)),
\label{eq:IK2}
\end{equation}
\begin{equation}
\theta_1 = \text{arctan2}(z, x) - \text{arctan2}(l_T \sin \theta_2, l_F + l_T \cos \theta_2),
\label{eq:IK3}
\end{equation}
where $x$, $z$ are the coordinates of the generated step trajectory as shown in Fig.~\ref{leg_traj} 

\begin{figure}[htbp]
\centering
\includegraphics[width=0.9\linewidth]{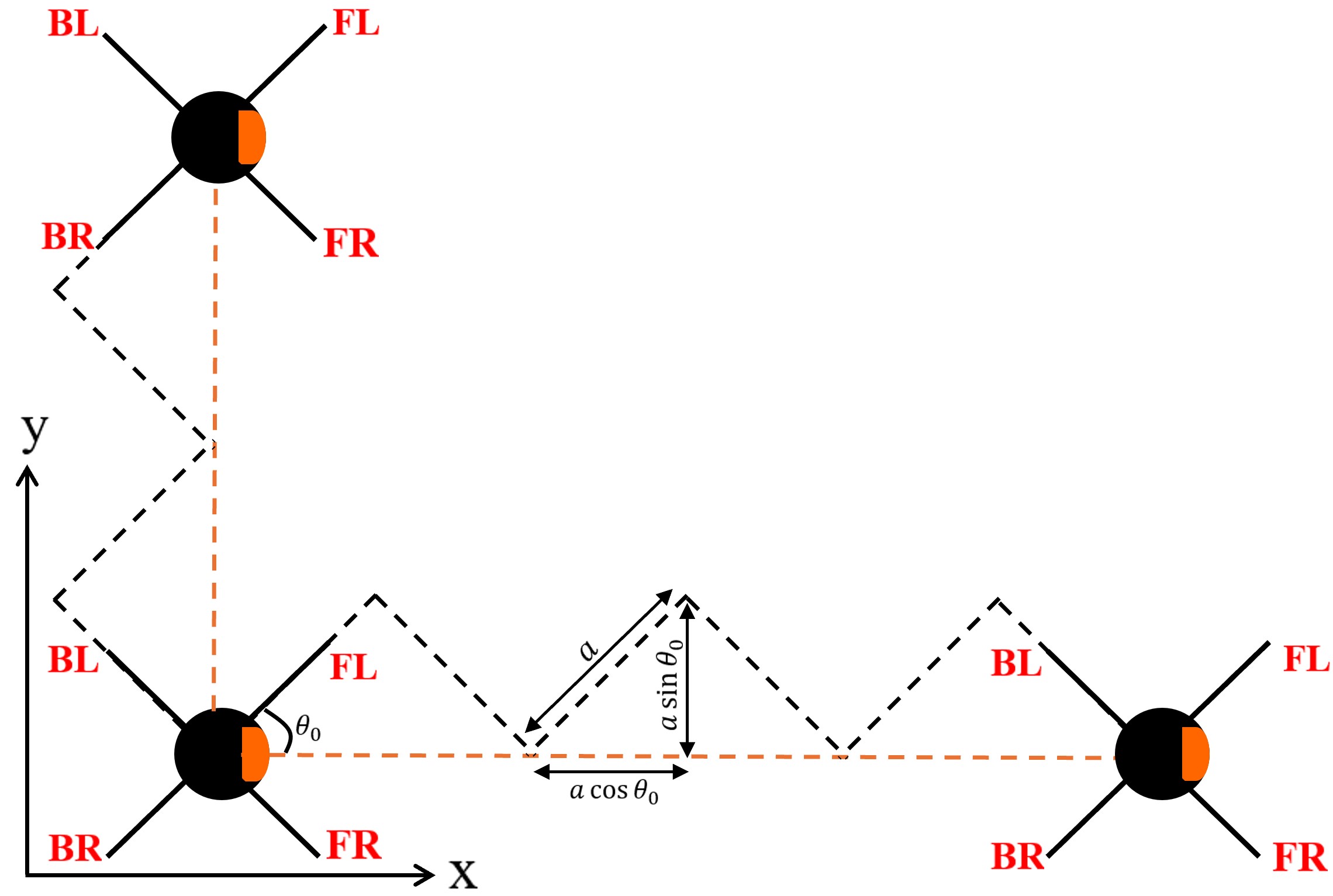} 
\caption{MorphoGear canter gait motion in the $x$ and $y$ directions.}
\label{locomotion}
\end{figure}

Due to the design constraints of MorphoGear, it can only locomote in the $x$ and $y$ directions (local body frame). In other words, it cannot traverse in the plane of any limb. Moreover, for this work, the shoulder angles $\theta_0$ for each limb have been kept constant. This results in the four limbs being equally spaced at 90{\textdegree} intervals throughout the canter gait motion. This is visualised in Fig.~\ref{locomotion}.

As described in \cite{b1}, to move from one point to another, the drone locomotes in a zigzag fashion. Each step can be broken down into two sub-steps. In the first sub-step, the forward left limb and the backward right limb will perform a step length of $a$ to move $a \cos(\theta)$ in the $x$ direction and $a \sin(\theta)$ in the $y$ direction. In the second sub-step, the forward right limb and the backward left limb will perform a step length of $a$ to move $a \cos(\theta)$ in the $x$ direction and $-a \sin(\theta)$ in the $y$ direction. In this manner, MorphoGear traverses in the $x$ direction. The same is repeated for locomotion in the $y$ direction with the $cos$ and $sin$ terms interchanged along with a change in the combination of the limbs. This method also allows the robot to locomote in any direction in the $xy$-plane without having to rotate.

The motion can be mathematically modelled as the following set of equations:
For $x$ direction we have:
\begin{equation}
\begin{pmatrix} x(k+1) \\ y(k+1) \end{pmatrix} = 
\begin{pmatrix} x(k) \\ y(k) \end{pmatrix} + 
\begin{pmatrix} a(k) \cos(\theta_0) \\ a(k) \sin(\theta_0) \end{pmatrix} + 
\begin{pmatrix} a(k) \cos(\theta_0) \\ -a(k) \sin(\theta_0) \end{pmatrix}
\label{eq:X}
\end{equation}

For $y$ direction we have:
\begin{equation}
\begin{pmatrix} x(k+1) \\ y(k+1) \end{pmatrix} = 
\begin{pmatrix} x(k) \\ y(k) \end{pmatrix} + 
\begin{pmatrix} a(k) \sin(\theta_0) \\ a(k) \cos(\theta_0) \end{pmatrix} + 
\begin{pmatrix} -a(k) \sin(\theta_0) \\ a(k) \cos(\theta_0) \end{pmatrix}
\label{eq:Y}
\end{equation}

\subsection{Model Predictive Control (MPC)}
For this study, an MPC controller was designed for the ground locomotion of the MorphoGear. Such a controller is mathematically formulated as below:
\begin{equation}
l(x,u) = \lVert x_u - x_r \rVert_Q^2 + \lVert u \rVert_R^2,
\label{eq:oldL}
\end{equation}
\begin{equation}
\min_{u} J(x,u) = \sum_{k=0}^{N-1} l(x_u(k), u(k)),
\end{equation}

subject to:
\[ x_u(k+1) = f(x_u(k), u(k)), \]
\[ x_u(0) = x_0, \]
\[ u(k) \in U, \quad \forall k \in [0, N-1], \]
\[ x(k) \in X, \quad \forall k \in [0, N] \]

The MPC is subject to the mathematical model of the system, which in this case is defined in Eqs.~\eqref{eq:X} and~\eqref{eq:Y}. Moreover, the MPC is also subject to the initial value along with constraints regarding inputs and states. For this case, the states are the $x$ and $y$ coordinates of MorphoGear, while the input is the step length $a$. The values of the $x$ and $y$ coordinates are governed by the mathematical model. Furthermore, the step length is constrained by MorphoGear's height and the full extension of its limbs. This is formulated as follows:
\begin{equation}
    a_{\text{max}} = \sqrt{(l_F + l_T)^2 - H^2}
    \label{eq:amax}
\end{equation}

The MPC's cost function consists of two quadratic costs. The first quadratic cost is associated with the deviation of the state vector $x$ weighted by the matrix $Q$. The second quadratic cost is associated with the deviation control input vector $u$ weighted by the matrix $R$. The weight $R$ is just a singular value rather than a matrix since there is only one input, the step length. For this study, no specific reference input value was taken; the control input itself is penalized without considering a reference value. The MPC parameters used in this framework are listed in Table~\ref{tab:mpcparams}.
\begin{table}[htbp]
    \centering
    \caption{MPC Parameters}
    \label{tab:mpcparams}
    \begin{tabular}{|l|c|}
    \hline
    Parameter & Value \\
    \hline
    Weight Matrix Q element, Qx & 3 \\
    Weight Matrix Q element, Qy & 3 \\
    Weight R & 0.2 \\
    Prediction Horizon & 3 \\
    \hline
    \end{tabular}
\end{table}

After calculation of the step length through MPC, the limb joint angles are found through the inverse kinematic Eqs.~\eqref{eq:IK1}, \eqref{eq:IK2} and \eqref{eq:IK3}. The required direction of motion is then checked, and the limb joint angles are then associated with the corresponding leg pairs as explained in Section  \ref{subsec:motion}.
\vspace{1pt}
\section{Analysis and Results}

\subsection{Experimental Setup}

\begin{figure}[htbp]
\centering
\includegraphics[width=0.8\linewidth]{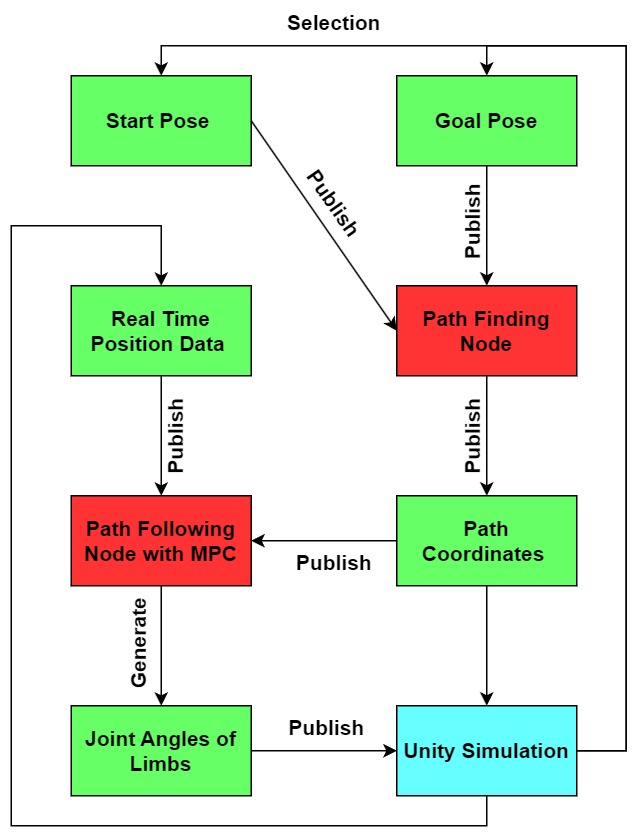} 
\caption{ROS communication between Unity 3D simulation environment and Path Planner with MPC for locomotion.}
\label{flowchart_sim}
\end{figure}

The algorithms were tested on a digital twin of MorphoGear in a Unity 3D environment. Robot Operating System (ROS) was used in order to subscribe and publish to corresponding topics for the digital twin. Fig.~\ref{flowchart_sim} shows this scheme.  

\subsection{Experimental Evaluation}
To evaluate the MPC, MorphoGear was given a straight trajectory from 0 to 3.6 m in the positive $x$-direction, which it traversed once with MPC and once without it (Fig.~\ref{error_graph}). After the validation of MPC through the previous experiment, MorphoGear was given a more complex trajectory to follow in the $xy$-plane (Fig.~\ref{mpc_graph}). Lastly, a mission was designed to test the path planning (ground and aerial) and following capabilities simultaneously (see Fig.~\ref{sim_setup}).

\subsection{Experimental Results}
\begin{figure}[h!]
\centering
\includegraphics[width=1\linewidth]{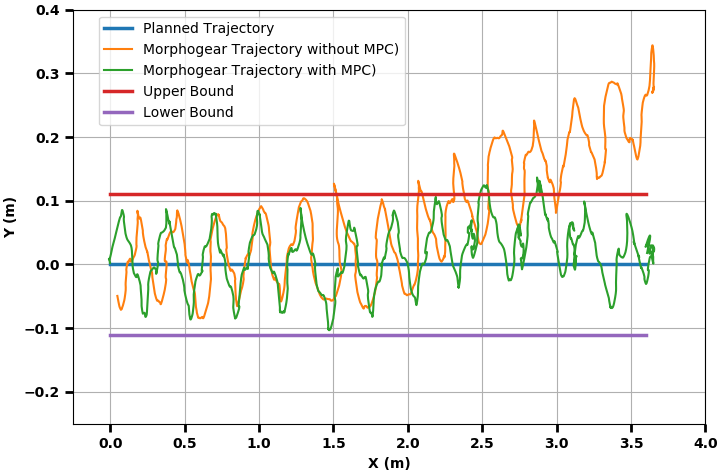} 
\caption{Experimental trajectories for MorphoGear with and without MPC.}
\label{error_graph}
\end{figure}

\begin{figure}[h!]
\centering
\includegraphics[width=0.9\linewidth]{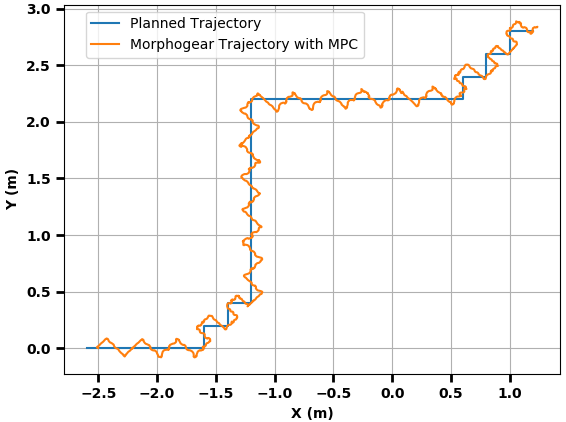} 
\caption{Experimental trajectories for MorphoGear with and without MPC for more complex trajectory.}
\label{mpc_graph}
\end{figure}

Fig.~\ref{error_graph} shows the results of MorphoGear's ability to follow a given path with and without MPC for a straight line trajectory of 3.6 m. Table~\ref{tab:errors} shows that the non-MPC trajectory had an RMSE of 9.77 cm with a maximum error of 22.58 cm, while the MPC trajectory had an RMSE of 0.91 cm with a maximum error of 1.85 cm. Additionally, it is also to be noted from Fig.~\ref{error_graph} that without any controller, once MorphoGear deviates from the given path, it does not tend to return back to it. On the other hand, with a closed-loop MPC system, MorphoGear not only follows the path but also stays within the maximum boundaries of the canter gait motion, which are governed by the maximum step length constraint in the MPC. Furthermore, the reduced step lengths computed by the controller can also be seen towards the end of the trajectory. A more complex trajectory is also shown in Fig.~\ref{mpc_graph} which the MorphoGear successfully followed.

\begin{table}[h!]
    \centering
    \caption{Path Following Errors}
    \label{tab:errors}
    \begin{tabular}{|l|c|c|}
        \hline
        & RMSE & Max Error\\ 
        \hline
        Without MPC & 9.77 cm & 22.58 cm\\ 
        \hline
        With MPC & 0.91 cm & 1.85 cm\\ 
        \hline
        Percentage Improvement with MPC & 90.68\% & 91.81\% \\
        \hline
    \end{tabular}
\end{table}

\begin{figure}[h!]
\centering
\includegraphics[width=1\linewidth]{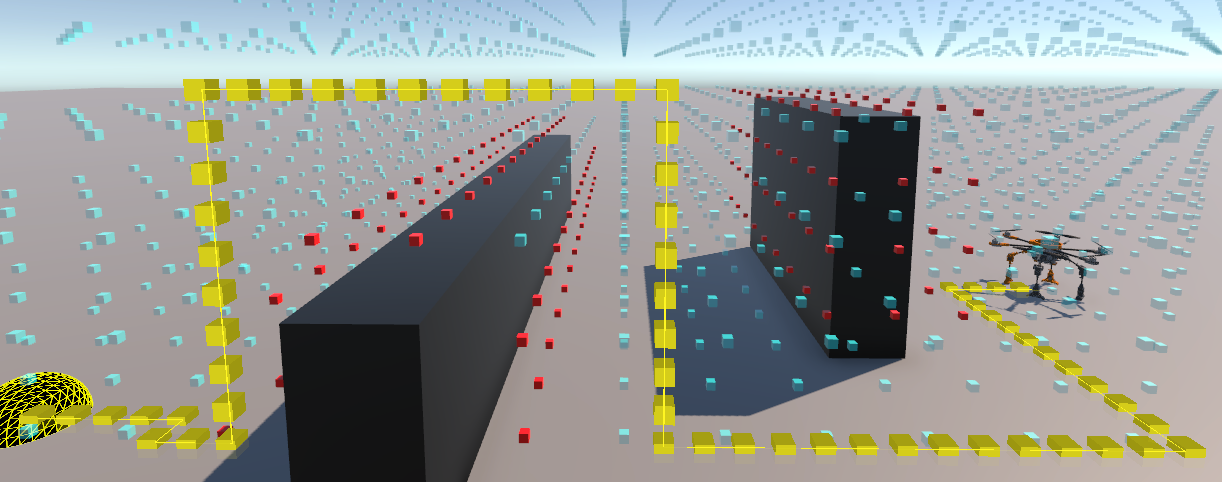} 
\caption{Unity Simulation Setup with the planned path (yellow line).}
\label{sim_setup}
\end{figure}

\begin{figure}[h!]
\centering
\includegraphics[width=0.6\linewidth]{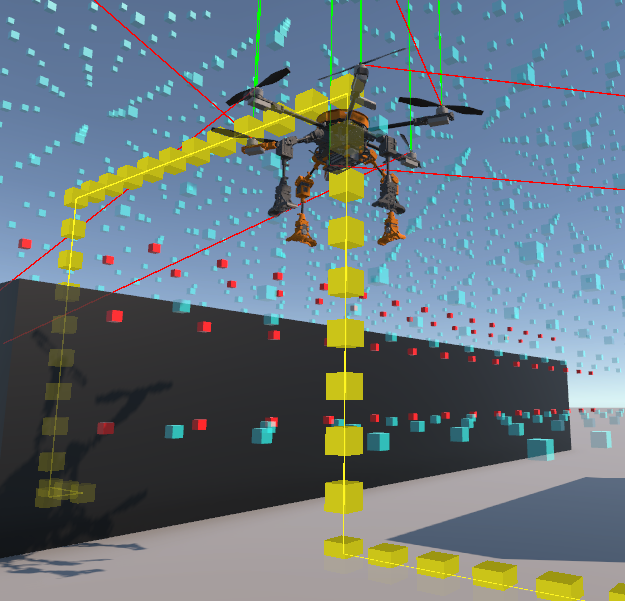} 
\caption{MorphoGear mid-flight in the Unity environment to overcome the obstacle in front of it.}
\label{sim_fly}
\end{figure}

Fig.~\ref{sim_setup} illustrates the path finding algorithm's ability to navigate in 2D where possible and switch to 3D where no ground path was possible. The MorphoGear followed the ground trajectory with the help of the MPC and then flew along the designated path and landed once the obstacle was cleared (see Fig.~\ref{sim_fly}). The results for this are plotted in Fig.~\ref{sim_setup}.

\begin{figure}[h!]
\centering
\includegraphics[width=1\linewidth]{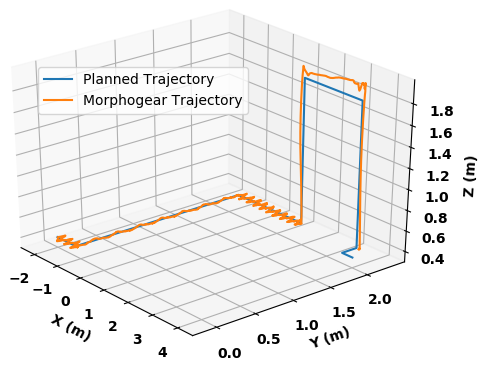} 
\caption{Trajectories plotted for the Unity simulation setup.}
\label{sim_graph}
\end{figure}
The experimental results showcase the significance of the developed algorithms ensuring effective path planning and path following, thus enhancing MorphoGear's capabilities for navigation in cluttered environments.

\section{Conclusion And Future Work}

 A bi-modal path planning algorithm that can switch between ground and aerial modes depending on the environment to leverage MorphoGear's dual nature has been developed. This was complemented with an MPC-based controller for the ground locomotion. The experimental results highlight the abilities of both algorithms in giving the desired trajectories between ground and aerial modes, and stable ground locomotion is achieved using an MPC-based path follower with minimum deviations (RMSE of 0.91 cm and maximum error of 1.85 cm). This showcases the ability of the MorphoGear to adhere to a given path and not stray from it. These developments portray adaptability in navigating complex environments, thus placing it as a pivotal tool in autonomous aerial and ground exploration.

In the future, we plan to develop the different path planning strategies for ground locomotion to reach optimal paths and sampling-based algorithms for aerial mode since they can potentially perform better in 3D scenarios and are relatively faster. In addition, we will also implement an MPC flight controller that takes into account MorphoGear's dynamics for stable flight that will provide stable flight and will enable the robot to use its limbs for manipulation during mid-flight.

\section*{Acknowledgements} 
Research reported in this publication was financially supported by the RSF grant No. 24-41-02039.

\addtolength{\textheight}{-12cm}   




\end{document}